\documentclass[runningheads]{llncs}

\usepackage{eccv}

\usepackage{eccvabbrv}

\usepackage{graphicx}
\usepackage{booktabs}
\usepackage{tabularx}

\usepackage{amsmath,amsfonts}
\usepackage{algorithmic}
\usepackage{algorithm}
\usepackage{array}
\usepackage{textcomp}
\usepackage{stfloats}
\usepackage{url}
\usepackage{verbatim}
\usepackage{graphicx}
\usepackage{cite}
\usepackage{url}
\usepackage{cite}
\usepackage{comment}
\usepackage{color}
\usepackage{graphicx,setspace}
\usepackage{multirow, array, booktabs,makecell,xspace,bm}
\usepackage{xcolor}
\usepackage{colortbl}

\usepackage{pgfplots} 
\usepackage{colortbl} 
\usepackage{booktabs}

\usepackage{pifont}
\usepackage{amsfonts}
\usepackage{arydshln}
\usepackage{dsfont}

\usepackage{stackengine}

\usepackage{scalerel}
\usepackage{esdiff} 

\usepackage{bbm}
\usepackage{colortbl}

\newcommand{\cmark}{\ding{51}}%
\newcommand{\xmark}{\ding{55}}%

\usepackage{hhline}
\usepackage{colortbl}

\usepackage{hyperref}

\usepackage{orcidlink}

\begin{document}

\title{Learning Trimodal Relation for Audio-Visual Question Answering with Missing Modality} 

\titlerunning{Learning Trimodal Relation for AVQA with Missing Modality}

\author{Kyu Ri Park\inst{1}\orcidlink{0009-0001-2140-9257} \and
Hong Joo Lee\inst{2,3}\thanks{Corresponding author}\orcidlink{0000-0001-6626-5683} \and
Jung Uk Kim\inst{1\dagger}\orcidlink{0000-0003-4533-4875}}

\authorrunning{K. Park et al.}

\institute{Kyung Hee University, Yong-in, South Korea \\
\email{\{kyuri0924, ju.kim\}@khu.ac.kr}\\ \and
Technical University of Munich, Munich, Germany,\\  \and
Munich Center for Machine Learning (MCML), Munich, Germany\\
\email{hongjoo.lee@tum.de}}


\maketitle

\begin{abstract}
  Recent Audio-Visual Question Answering (AVQA) methods rely on complete visual and audio input to answer questions accurately. However, in real-world scenarios, issues such as device malfunctions and data transmission errors frequently result in missing audio or visual modality. In such cases, existing AVQA methods suffer significant performance degradation. In this paper, we propose a framework that ensures robust AVQA performance even when a modality is missing. First, we propose a Relation-aware Missing Modal (RMM) generator with Relation-aware Missing Modal Recalling (RMMR) loss to enhance the ability of the generator to recall missing modal information by understanding the relationships and context among the available modalities. Second, we design an Audio-Visual Relation-aware (AVR) diffusion model with Audio-Visual Enhancing (AVE) loss to further enhance audio-visual features by leveraging the relationships and shared cues between the audio-visual modalities. As a result, our method can provide accurate answers by effectively utilizing available information even when input modalities are missing. We believe our method holds potential applications not only in AVQA research but also in various multi-modal scenarios. The code is available at \href{https://github.com/VisualAIKHU/Missing-AVQA}{https://github.com/VisualAIKHU/Missing-AVQA}.
\end{abstract}

\section{Introduction}
\label{sec:intro}
In the era of artificial intelligence, research efforts aimed at understanding scenes by integrating multi-modal information have made significant progress. A notable example in this field is Audio-Visual Question Answering (AVQA), which integrates video, audio, and text inputs to comprehend complex situations and generate responses by assimilating relevant video and audio information based on the questions. It involves extracting salient information from inputs and training networks to identify correlated features for accurate prediction.

Due to the importance of AVQA, many recent studies have been proposed. Li et al. \cite{Music-AVQA} introduced a large dataset named MUSIC-AVQA \cite{Music-AVQA} and a spatio-temporal based audio-visual network tailored for the AVQA task. Yun et al. introduced Pano-AVQA \cite{Pano-AVQA} that focuses on AVQA within panoramic videos. Pano-AVQA established a new benchmark comprising spherical spatial relation QAs and audio-visual relation QAs. Transformer-based models trained on Pano-AVQA exhibit an enhanced semantic understanding of panoramic surroundings. Yang et al. proposed HAVF\cite{2022avqa}, a study focused on developing a hierarchical audio-visual fusing module to model semantic correlations among audio, visual, and text modalities. Li et al. proposed a Progressive Spatio-Temporal Perception Network (PSTP-Net)\cite{PSTP} that progressively identifies key spatio-temporal regions relevant to a question.

\begin{figure}[t]
    \begin{minipage}[b]{1.0\linewidth}
	\centering
        \centerline{\includegraphics[width=12.4cm]{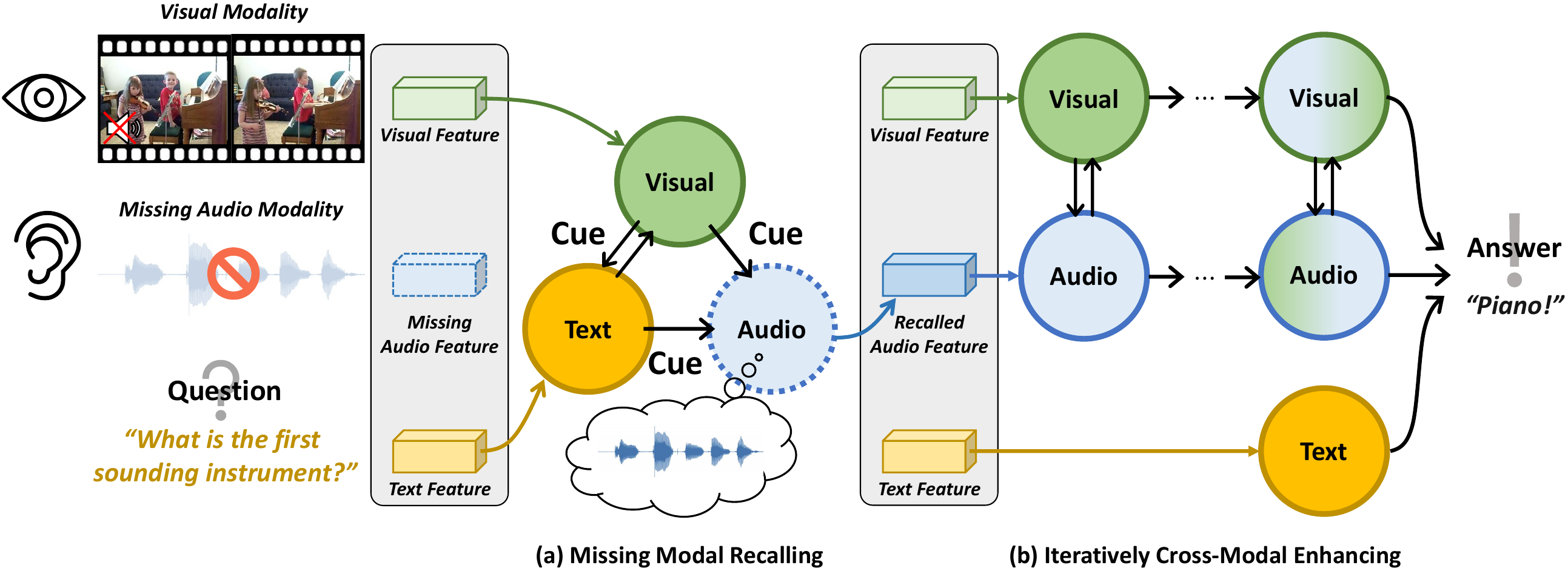}}
        \end{minipage}
        \vspace{-0.5cm}
	\caption{Concept diagram of our methodology. Leveraging mutual cues in trimodal relations to recall and enhance missing information.}
    \vspace{-0.5cm}
    \label{fig:1}
\end{figure}

Although aforementioned methods have led to significant improvements, a critical issue remains unaddressed: the inherent reliance of existing AVQA approaches on complete input modalities. In real-world scenarios, it is common for certain modalities to be unavailable due to device malfunctions\cite{missingmodal_decoding,missingmodal_ssm} or environmental factors\cite{missingmodal,missingmodal_av,missing_adversarial,Ma_2022_CVPR}, such as low-light conditions or noisy surroundings. Therefore, the limitations of current AVQA methods become evident when dealing with incomplete input modalities, leading to reduced performance and inaccurate responses. Some previous works tried to handle missing modality issues by generating pseudo features\cite{missing_prompt,Shaspec,missing_AAAI}. However, these approaches face challenges when applied to the AVQA task due to its unique complexities. Existing methods have primarily tackled the problem by handling modality in one-to-one pairs\cite{Ma_2022_CVPR}, neglecting the inter-dependencies among different modalities. Consequently, they can only generate pseudo features corresponding to the given modality, failing to consider the broader multi-modal context. However, the AVQA task poses a more intricate challenge, necessitating visual-audio reasoning\cite{av1, av2, av3, av4} across diverse contexts to answer complex questions effectively. AVQA requires a nuanced understanding of the context of question to generate relevant information for accurate responses, making it essential to flexibly generate pseudo features\cite{Shaspec} for missing modalities depending on the specific question.

In this paper, we propose a new approach with a novel AVQA framework to address the issue of missing modalities. This approach is based on analyzing human cognitive psychology\cite{human1,human2,human3}, specifically the ability to recall information through audio-visual integration\cite{kim2023enabling,raij2000audiovisual,kim2024learning,um2023audio,calvert2001detection}. \cref{fig:1} shows the conceptual diagram of our method. When only one modality is available (\textit{e.g.,} visual), humans can retrieve associated information of another modality (\textit{e.g.,} the sound of a piano) through the integration of visual-text cues. We propose a Relation-aware Missing Modal (RMM) generator. This generator takes visual-text or audio-text information as input. By associating the available modalities, the RMM generator recalls the missing modality and derives a pseudo feature for it. By determining the correlation between the two modalities, we can effectively identify the missing modality.

Next, we introduce an Audio-Visual Relation-aware (AVR) diffusion model to enhance the features generated by the RMM generator with mutual cues. The real feature of the available modality (visual) and the pseudo feature of the missing modality (audio) are combined and passed into the AVR diffusion model. This process enhances the features by leveraging mutual cues from the different modalities. Then, the output of the diffusion becomes the input for the AVQA predictor in place of the missing modality. These two steps are designed with new insights to mimic the two steps of human trimodal interpretation. The RMM uses trimodal relations for missing modalities, and the AVR enhances features using complementary audio-visual relations. This approach ensures that the enhanced features from both modalities contribute to more accurate and robust predictions, improving the overall performance of the AVQA system. 

The main contributions of this paper are summarized as follows:
\begin{itemize}
    \item We introduce a novel AVQA framework that simultaneously addresses the issue of missing modalities.
    \item We propose the Relation-aware Missing Modal (RMM) generator, a new approach to recall the information of the missing modality by associating two existing modalities and deriving a pseudo feature.
    \item We present the Audio-Visual Relation-aware (AVR) diffusion model, which emphasizes the mutual enhancement of modalities by effectively utilizing and referencing information from each other. 
\end{itemize}

\section{Related Work}
\subsection{Audio-Visual Question Answering}
Recently, several works have used audio, visual, and text modalities for multi-modal scene understanding. Schwartz et al.\cite{AVSD} proposed a baseline for audio-visual scene understanding, consisting of feature extractors, a multi-modal attention module, and an answer generation module. Yun et al. introduced the Pano-AVQA network\cite{Pano-AVQA} for semantic scene understanding in panoramic videos, proposing spherical spatial embedding methods and using equirectangular and NFoV\cite{NoFV} projections to reduce visual distortion during feature extraction. Li et al.\cite{Music-AVQA} developed the MUSIC-AVQA dataset for AVQA tasks in musical performance scenes, offering more complex relations such as existential, location, counting, comparative, and temporal information, and proposed spatial and temporal grounding networks. Li et al.\cite{PSTP} also proposed PSTP-Net, which finds key spatio-temporal regions from video using a temporal segment selection module, spatial regions selection module, and audio-guided visual attention module.

Most of these works assumed the completeness of modality which may not considered missing modality. In real-world situations, some modalities can be missed due to device malfunctions or environmental constraints\cite{missingmodal,missingmodal_av,missing_adversarial,Ma_2022_CVPR}. Under these missing modalities situations, AVQA systems can become unstable  and fail to work properly. Unlike previous works, this paper focuses on handling the problem of missing modalities.

\subsection{Missing Modality in Multi-modal Learning}
Recently, some works have addressed the missing modality problem in multi-modal learning \cite{kim2022towards,missing_AAAI,lee2022audio,missing_prompt,Shaspec,lee2022weakly}. Woo et al. \cite{missing_AAAI} investigated the effects of architecture, data augmentation, and regularization under missing modalities and proposed an Action Masked Auto Encoder (ActionMAE) that generates pseudo features of missing modalities for inference. Wang et al. proposed Shared-Specific Feature Modeling (ShaSpec)\cite{Shaspec}, which extracts shared and modality-specific features to enhance input data representation using shared and specific encoders. Lee et al. \cite{missing_prompt} introduced a missing-aware prompting method for transformer models that helps the model be aware of the missing modality by attaching missing-aware prompts at the input. Woo et al. \cite{missing_AAAI} also proposed a transformer architecture to fuse features from all modalities into a comprehensive set using hybrid modality-specific encoders, intra-modal transformers, and inter-modal transformers.
In the context of autonomous driving, Choi et al.\cite{driving1} proposed a Shared Cross-modal Embedding method to encode features effectively, addressing missing modality issues. Additionally, Wu et al.\cite{driving2} addressed the missing modality problem using a knowledge distillation method with a vision teacher, an auditory teacher, and an audio-visual student.
These studies have demonstrated significant success in addressing the missing modality problem. However, their applicability to the AVQA task is limited due to its unique complexities. Existing methods primarily handle modality in one-to-one pairs, overlooking the interdependencies among different modalities. For example, \cite{missing_AAAI} pairs input images, depth images, and IR images to generate estimated features. In contrast, the AVQA task demands a nuanced understanding of the question context to produce relevant information, necessitating the flexible generation of pseudo features for missing modalities based on the specific question. Consequently, this paper proposes a method to tackle the missing modality problem more effectively by comprehending the context of questions and generating missing modality information that corresponds to the context of the questions.

\subsection{Diffusion Based Models}
The diffusion model generates data through an iterative process of adding and removing noise. It consists of a forward process that adds noise at each time step and a reverse process that removes noise at each time step. Diffusion models represent a promising group of generative models that simplify the data generation process into a step-by-step noise reduction technique\cite{DDPM}. Diffusion models have been shown to perform well in several image generation tasks\cite{dhariwal2021diffusion}, including super-resolution\cite{qiu2022learning, saharia2022image,zeng2021improving}, effective image restoration\cite{kawar2022denoising}, image processing\cite{kim2021diffusionclip, meng2021sdedit}, text conditioning\cite{gu2022vector,nichol2021glide,rombach2022high}, image inpainting\cite{saharia2022palette}, and more. Stable diffusion\cite{rombach2022high} built a DPM (Diffusion Probabilistic Model) in latent space to reduce the number of pixels. Most of these previous works on diffusion take only one modality as input to the diffusion process and derive the output of the same modality. In contrast, our AVR diffusion model integrates audio and visual modalities simultaneously, ensuring effective fusion and thorough learning of the diffusion process for enhanced feature extraction.

\section{Methodology}

\begin{figure}[t]
    \centering
        \centerline{\includegraphics[width=12.5cm]{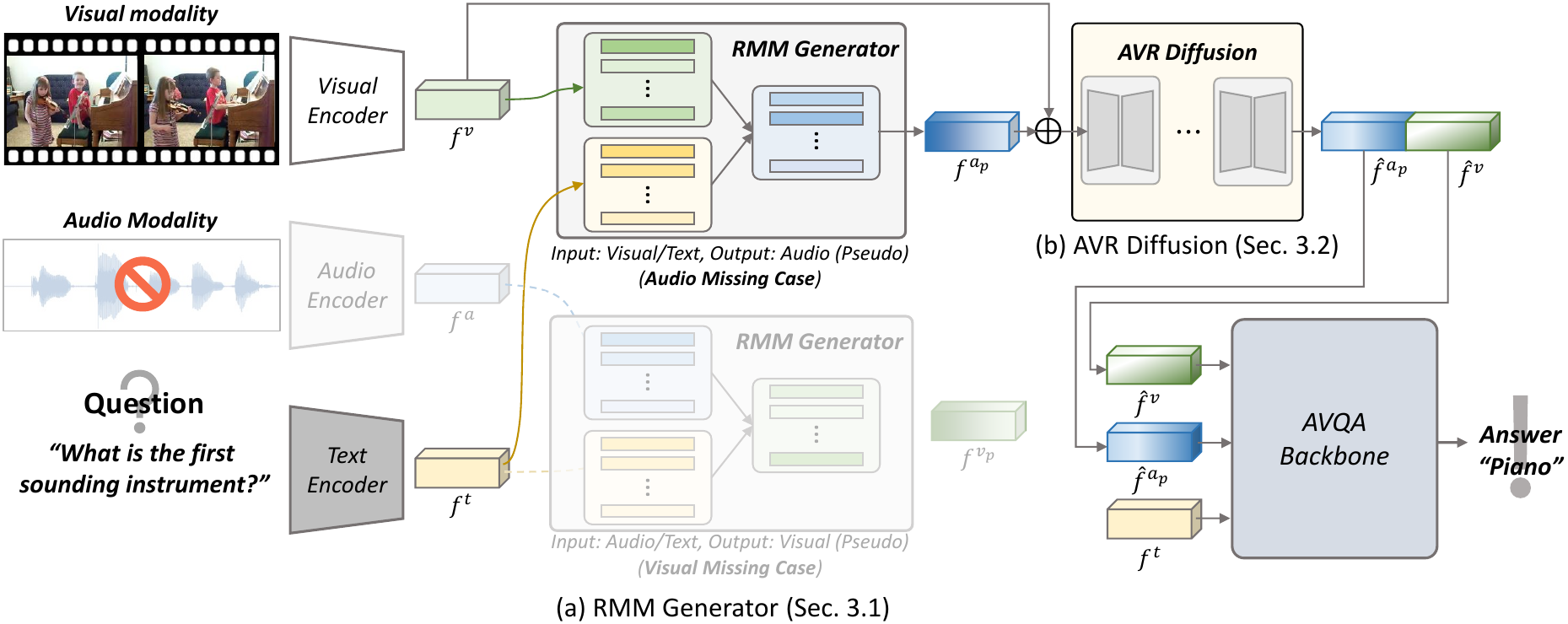}}
	\caption{Overall architecture of the proposed AVQA framework for missing modality (audio missing example). We introduce (a) Relation-aware Missing Modal (RMM) generator and (b) Audio-Visual Relation-aware (AVR) diffusion model. More details for learning RMM generator and AVR diffusion are in Sec. 3.1 and Sec. 3.2.}
    \vspace{-0.5cm}
    \label{fig:2}
\end{figure}

\cref{fig:2} shows the overall architecture of the proposed AVQA framework addressing missing modalities during inference. The visual modal input, audio input, and question pass through their corresponding encoders to obtain ${f}^{v}$, ${f}^{a}$, ${f}^{t}$. Our method consists of three major components: (1) Relation-aware Missing Modal (RMM) generator (see \cref{fig:2} (a)), (2) Audio-Visual Relation-aware (AVR) diffusion model (see \cref{fig:2} (b)), and (3) AVQA backbone. For example, in the case of audio being missing, the RMM generator generates a pseudo audio feature for the missing modality (\textit{e.g.,} audio) using the existing modalities (\textit{e.g.,} visual and question). Since our work addresses missing scenarios during the inference phase, the RMM generator is trained to learn pseudo audio features that closely resemble the real audio features. Then, with the pseudo audio feature and the visual modal feature, the AVR diffusion model enhances the feature representation of each modality by referring to cross-modal knowledge. To train the AVR diffusion model, real features ${f}^{v}$ and ${f}^{a}$ are utilized during training. Finally, the features of the two modalities obtained through the AVR diffusion model, along with the question feature, are passed through the AVQA backbone to estimate the answer. A similar process is employed to account for the missing visual modality. 

Our novelty consists of two key steps: (1) using trimodal relations to handle missing modalities and (2) enhancing features with complementary audio-visual relations. The RMM generator leverages trimodal relations to generate pseudo features for the missing modality. Then, the AVR diffusion model uses cross-modal knowledge to enhance the feature representation of each modality, combining the pseudo audio feature with the visual feature.
By doing so, even when a single-modal input is missing during the inference phase, our RMM generator and AVR diffusion model can recall the missing modal information and improve feature representation, showing robust performance for AVQA.
\vspace{-0.3cm}
\begin{figure}[t]

	\centering
        \centerline{\includegraphics[width=12.5cm]{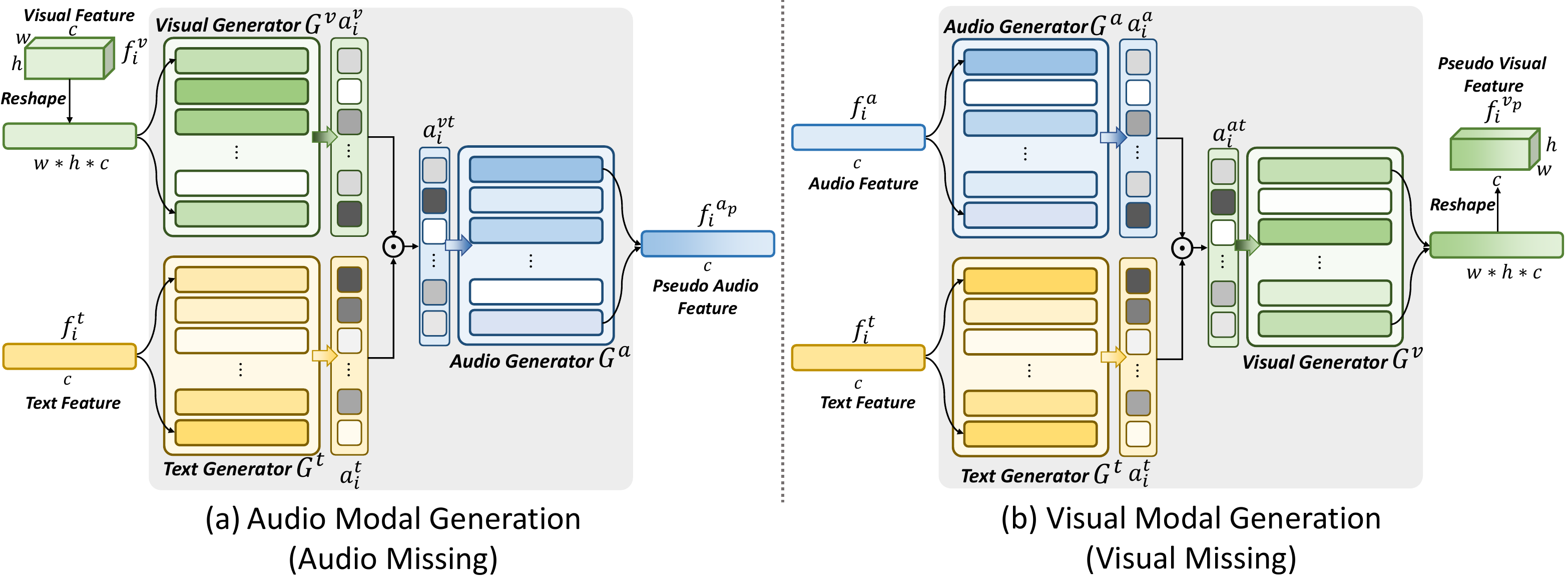}}
        \vspace{-0.3cm}

	\caption{RMM generator operates in two scenarios: (a) generating pseudo audio when audio input is missing, and (b) generating pseudo visuals when visual input is missing. Based on the addressing vector (\textit{i.e.,} ${a}^{vt}_i$, ${a}^{at}_i$), pseudo modality feature is obtained by a weight summation with the corresponding generator. Each generator in the three modalities shares weights.}
    \label{fig:3}
    \vspace{-0.5cm}
\end{figure}

\subsection{Relation-aware Missing Modal Generator}
The Relation-aware Missing Modal (RMM) generator is inspired by the remarkable ability of human brain to integrate and process different types of information from various sensory modalities. When humans encounter content, they can often infer missing audio information by analyzing visual cues and context, and similarly, infer visual information from audio cues. This capability arises from shared cues between different modalities that come from the same object or scene. For instance, when only auditory information of piano sound is provided, humans can perceive the sound and recall the image of a piano (visual information). Conversely, when watching a muted video of a piano performance, humans can recall the piano sound (auditory information).

The RMM generator mimics this cognitive mechanism by leveraging the relationships between visual, auditory, and textual modality to effectively handle multi-modal inputs. It uses a slot-based architecture, where each modality is represented by $L$ learnable parameter vectors that capture essential features. These slots are crucial for recalling and reconstructing missing information. By analyzing spatio-temporal patterns in the input modality, the RMM generator identifies correlations between modalities, enabling it to predict and restore missing information precisely. This approach mirrors the interconnected cortical regions of brain specialized in visual, auditory, and language processing, enhancing the ability of model to synthesize and interpret diverse modality\cite{lahat2015multimodal}.

\cref{fig:3} (a) illustrates the scenario when the audio modality is missing\cite{missingmodal_av}. The visual feature $f^v_i$ and the text feature $f^t_i$ are passed through RMM generator. Particularly, $f^v_i$ is flattened to create a vector  $\bar{f}^v_i\in\mathbb{R}^{1\times whc}$. RMM generator is composed of three modal generators. Each generators consists of $L$ slots denoted as $\textbf{G}=\{\textbf{G}^v_j, \textbf{G}^a_j, \textbf{G}^t_j\}$$_{j=1}^L$ $(\textbf{G}^v_j\in\mathbb{R}^{1\times whc}, \textbf{G}^a_j, \textbf{G}^t_j\in\mathbb{R}^{1\times c})$. These slots represent the number of learnable parameter vectors for each modality in the RMM generator, facilitating the generation of missing features by establishing relationships between modalities. Each slot contains data that serves as a reminder of the corresponding modality. 
When the audio modality is missing, the visual generator $\textbf{G}^{v}$ and text generator $\textbf{G}^{t}$ use  $\bar{f}^v_i$ and $f^t_i$ to determine how much of the $L$ slots from the auditory generator $\textbf{G}^{a}$ are needed to recall the audio information. That is, $\bar{f}^v_i$ and $L$ slots of $\textbf{G}^{v}$ are calculated to obtain visual addressing vector ${a}^{v}_i=\{a^v_{i1},\dots, a^v_{iL}\}\in \mathbb{R}^{1\times L}$. Each element of ${a}^{v}_i$ is calculated as:
\vspace{-0.2cm}
\begin{equation}
	\begin{gathered}
		a^v_{ij}=\frac{\exp{(s^v_{ij})}}{\sum_{m=1}^{L}\exp{(s^v_{im})}},\,\, s^v_{ij}=\frac{\bar{f}^v_i\cdot \textbf{G}_{j}^{v\,\top}}{\sqrt{\,c\,\,}}.
	\end{gathered}
	\label{eq:1}
\end{equation}
A high value of $a^v_{ij}$ means a strong correlation between $\bar{f}^{v}_{i}$ and $\textbf{G}^v_{j}$. Else, they are weakly correlated. Similar to \cref{eq:1}, text addressing vector ${a}^{t}_i=\{a^t_{i1},\dots, a^t_{iL}\}\in \mathbb{R}^{1\times L}$ is obtained using $f^t_i$ and $\textbf{G}^{t}$. Next, visual-text addressing vector ${a}^{vt}_i$ is computed as ${a}^{vt}_i=softmax({a}^{v}_i\circ{a}^{t}_i)$ ($\circ$ is element-wise multiplication). We designed ${a}^{vt}_i$ to highlight the strongly correlated audio slots among $L$ slots by associating visual and textual modalities. Given ${a}^{v}_i$ and ${a}^{t}_i$ (each vector represents the association between each modal feature and $L$ slots), we perform element-wise multiplication and softmax to generate ${a}^{vt}_i$. Then, ${a}^{vt}_i$ and $\textbf{G}^{a}$ are aggregated to generate pseudo audio feature ${f}^{{a}_{p}}_i\in\mathbb{R}^{1\times c}$, which can be represented as:
\vspace{-0.2cm}
\begin{equation}
    {f}^{{a}_{p}}_i=\sum_{j=1}^{L}a^{vt}_{ij} \cdot \textbf{G}^a_{j}.
\end{equation}
If the $j$-th element of $a^{vt}_{ij}$ is high, the $j$-th slot of $\textbf{G}^a_{j}$ will play a more important role in recalling the audio modal information. Likewise, \cref{fig:3} (b) illustrates the case where the visual modality is missing. The RMM generator accepts an audio feature \( f^{a}_{i} \) and a text feature \( f^{t}_{i} \) to produce a pseudo visual feature \( {f}^{{v}_{p}}_i \) in \( \mathbb{R}^{1 \times w \times h \times c} \). This process is similar to that shown in \cref{fig:3} (a). 

It is important to note that both scenarios, audio missing (\cref{fig:3}(a)) and visual missing (\cref{fig:3} (b)), are considered during the training phase. Therefore, the weight parameters of each \( \textbf{G}^v \), \( \textbf{G}^a \), and \( \textbf{G}^t \) in \cref{fig:3} (a) and (b) are shared. Through the aforementioned process, the RMM generator is able to generate the missing modality feature by associating the text with a remaining modality. The generated pseudo feature can takes the place of the missing modality.
\vspace{-0.3cm}

\subsubsection{Relation-aware Missing Modal Recalling Loss.}
The main purpose of designing our RMM generator is to effectively recall the missing modal information during inference. Therefore, we propose Relation-aware Missing Modal Recalling (RMMR) loss $\mathcal{L}_{rmmr}$ to guide the outputs of the RMM generator to closely resemble the actual modal features. $\mathcal{L}_{rmmr}$ consists of the two losses: audio recalling loss $\mathcal{L}_{a}$ for audio modal missing, visual recalling loss $\mathcal{L}_{v}$ for visual modal missing. $\mathcal{L}_{a}$ guides the pseudo audio feature ${f}^{{a}_{p}}_i$ from RMM generator to link the semantic knowledge with the real audio feature ${f}_{i}^{a}$. Likewise, $\mathcal{L}_{v}$ guides the pseudo visual feature ${f}^{{v}_{p}}_i$ to be similar to the real visual feature ${f}_{i}^{v}$. The $\mathcal{L}_{a}$ and $\mathcal{L}_{v}$ are formulated as follows: 
\vspace{-0.2cm}
\begin{equation}
\begin{gathered}
\mathcal{L}_{a}=\frac{1}{N}\sum_{i=1}^{N}\left \| f^{a}_{i} - {f}^{{a}_{p}}_i \right \|_2^2,\,\,\,
\mathcal{L}_{v}=\frac{1}{N}\sum_{i=1}^{N}\left \| f^{v}_{i} - {f}^{{v}_{p}}_i \right \|_2^2,
\end{gathered}
\label{eq:3}
\end{equation}
where $N$ indicates the number of the batch size.

Finally, we propose Relation-aware Missing Modal Recalling (RMMR) loss $\mathcal{L}_{rmmr}$. It can be represented as:
\begin{equation}
\begin{gathered}
\mathcal{L}_{rmmr}=\mathcal{L}_{a} + \mathcal{L}_{v}.
\end{gathered}
\label{eq:4}
\end{equation}
Through \cref{eq:4}, our AVQA framework with RMM generator can effectively recall the missing modal information. As a result, our AVQA method can provide accurate answers to questions, even in situations where one of the modalities is missing during the inference phase.
\vspace{-0.3cm}

\begin{figure}[t]
    \begin{minipage}[b]{1.0\linewidth}
	\centering
        \centerline{\includegraphics[width=12.4cm]{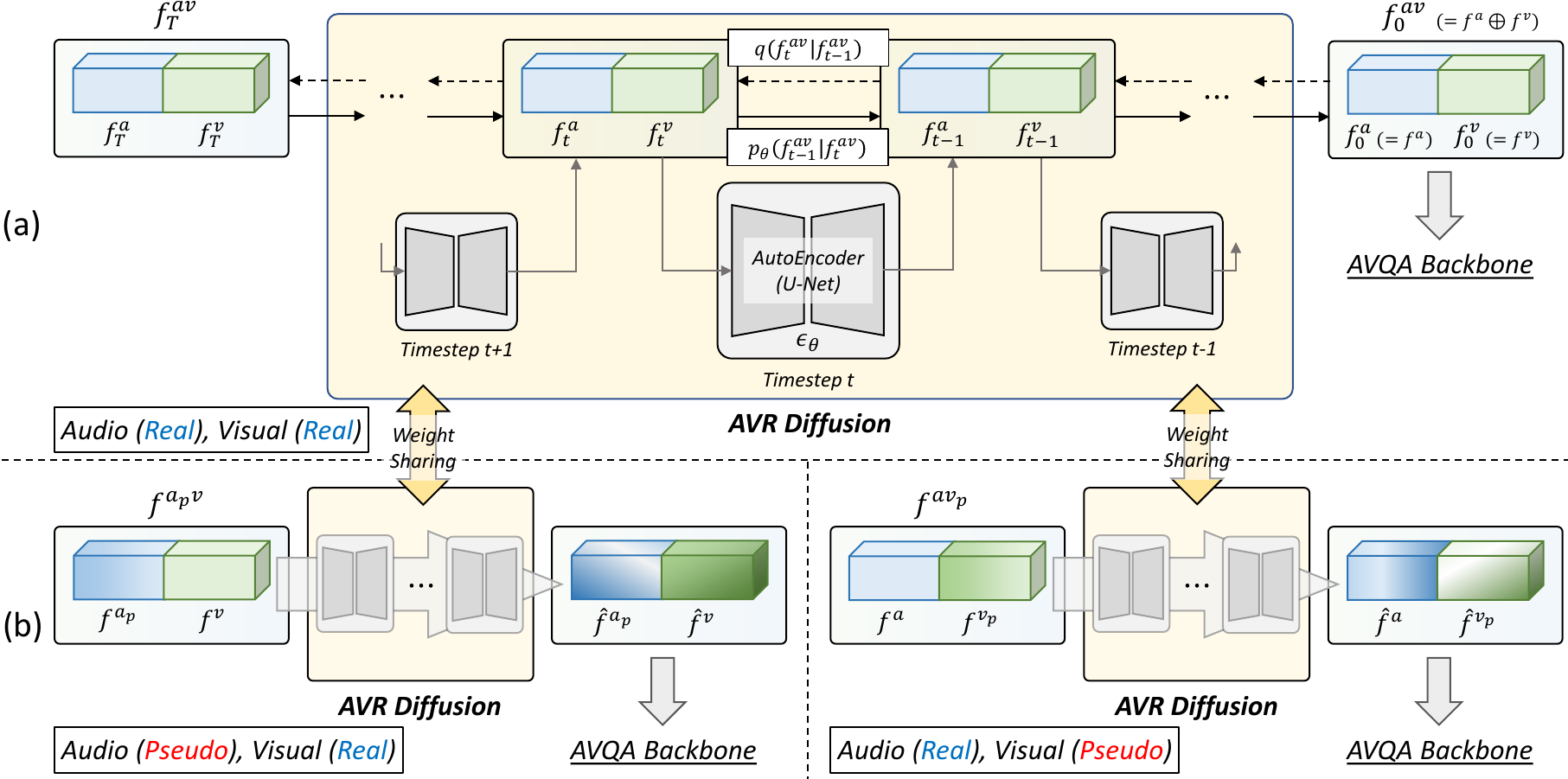}}
        \end{minipage}
	\caption{AVR diffusion process illustrates how the model learns to generate enhanced features for both audio and visual modalities by leveraging cross-modal knowledge. (a) depicts the diffusion and reverse process of concatenated features between real features, while (b) represents the reverse process of features where the pseudo feature and the real feature are concatenated.}
    \vspace{-0.2cm}
    \label{fig:4}
\end{figure}

\subsection{Audio-Visual Relation-aware Diffusion Model}

In this section, we introduce the proposed Audio-Visual Relation-aware (AVR) diffusion model. The goal of this model is to enhance the feature representation of both the missing modality (\textit{e.g.,} audio) as well as the original counterpart modality (\textit{e.g.,} visual). As illustrated in \cref{fig:4} (a), the process begins by combining the real audio feature (\textit{i.e.,} $ {f}^{a} $) and the visual feature (\textit{i.e.,} $ {f}^{v} $) through concatenation, resulting in the combined feature (\textit{i.e.,} $ {f}^{av} $). This combined feature is then passed through a diffusion process, which includes a forward process $ q $ (adding noise) and a reverse process $ p_{\theta} $. In the reverse process, $ p_{\theta} $ estimates the steps of the forward process in reverse, using the weight parameter $ \theta $ of the autoencoder \cite{DDPM}. This allows the AVR diffusion model to learn how to recover the original data from the noise effectively. The forward process $q$ and reverse process $p_{\theta}$ at time step $t\in[0, T]$ is defined as:

\begin{equation}
\begin{gathered}
    {q}({f}^{av}_{t}|{f}^{av}_{t-1}) = \mathcal{N}
    ({f}^{av}_t;\sqrt{1-\beta_{t}}{f}^{av}_{t-1},\beta_{t}I), \,\,\, t \in [1,T],
\end{gathered}
\label{eq:5}
\end{equation}
\begin{equation}
\begin{gathered}
{p}_{\theta}({f}^{av}_{t-1}|{f}^{av}_{t}) = \mathcal{N}({f}^{av}_{t-1};\mu_{\theta}({f}^{av}_{t},t), \Sigma_{\theta}({f}^{av}_{t},t)),
\end{gathered}
\label{eq:6}
\end{equation}
where $\beta_{t}$ indicates hyper-parameters to control amount of noise at time step $t$, $\mu_{\theta}({f}^{av}_{t},t)$ and $\Sigma_{\theta}({f}^{av}_{t},t))$ denote mean and variance, respectively. In this process, combining features of audio and visual modalities enables mutual information utilization, resulting in enhanced feature representations. This process is repeated $T$ timesteps. As a result, our AVR diffusion has learned the ability to enhance feature representations for a given input. 

Since we aim to address the missing modalities (audio or visual) in the inference phase, we also combine (${f}^{{a}_{p}}$, ${f}^{v}$) to generate ${f}^{{{a}_{p}}{{v}}}$ for audio missing (see \cref{fig:4} (b) left) and (${f}^{a}$, ${f}^{{v}_{p}}$) to generate ${f}^{{{a}}{{v}_{p}}}$ for visual missing (see \cref{fig:4} (b) right). ${f}^{{a_p v}}$ and ${f}^{{a v_p}}$ go through the reverse process of AVR diffusion to produce $\hat{f}^{{a_p v}}$ and $\hat{f}^{{a v_p}}$. Finally, after leveraging cross-modal knowledge through AVR diffusion, the audio and visual features are separated, \textit{i.e.,} ($\hat{f}^{a}$, $\hat{f}^{{v}_{p}}$) and ($\hat{f}^{{a}_{p}}$, $\hat{f}^{v}$) to use each individual feature as an input to the AVQA backbone. Note that the existing AVQA networks \cite{AVSD,Music-AVQA,Pano-AVQA,PSTP} require individual inputs (audio, visual, question (text)) to answer the questions.

The role of our AVR diffusion can be highlighted as follows: (1) Through the diffusion process, AVR diffusion learns the ability to generate the enhanced features for both audio-visual modalities by jointly leveraging cross-modal knowledge of ${f}^{av}$. (2) Also, the features of the pseudo modality and the counterpart original modality are combined and passed through AVR diffusion to further enhance their representations, considering the missing case in the inference phase.
\subsubsection{Audio-Visual Enhancing Loss.}
To guide AVR diffusion can perform the aforementioned roles, we introduce the Audio-Visual Enhancing (AVE) loss $\mathcal{L}_{ave}$, which can be represented as:
\begin{equation}
    \begin{gathered}
    \mathcal{L}_{ave} = \mathbb{E}_{\epsilon\sim\mathcal{N}(0,I)}
    [||\hat{\epsilon}_{\theta}({f}^{{av}}_{t},{t}) - \epsilon \|_2^2],
    \end{gathered}
\end{equation}
where $\epsilon$ denote noise in normal distribution $\mathcal{N}(0,I)$ and $\hat{\epsilon}_{\theta}({f}^{av}_{t},{t})$ denotes the prediction of the autoencoder of AVR diffusion at the $t$-th time step.

\subsection{Total Loss}
To learn the AVQA network to perform robustly in missing modality scenarios, the total loss function is defined as follows:

\begin{equation}
    \begin{gathered}
        \mathcal{L}_{avqa} = \mathcal{L}_{ce}({f}^{a}, {f}^{v}, {f}^{t}) + \mathcal{L}_{ce}(\hat{f}^{{a}_p}, \hat{f}^{v}, {f}^{t}) + \mathcal{L}_{ce}(\hat{f}^{a}, \hat{f}^{{v}_p}, {f}^{t}),\\
        \mathcal{L}_{Total} = \mathcal{L}_{avqa} + \lambda_1\mathcal{L}_{rmmr} + \lambda_2\mathcal{L}_{ave},
    \end{gathered}
    \label{eq:8}
\end{equation}
where $\lambda_1$,$\lambda_2$ denote the balancing hyper-parameters, $\mathcal{L}_{avqa}$ denotes the loss function of the AVQA predictors \cite{Music-AVQA,Pano-AVQA,AVSD,PSTP}, and $\mathcal{L}_{ce}$ indicates the cross-entropy loss when (${f}^{a},{f}^{v},{f}^{t}$), ($\hat{f}^{{a}_p}, \hat{f}^{v}, {f}^{t}$) and ($\hat{f}^{a}, \hat{f}^{{v}_p}, {f}^{t}$) pairs are input to AVQA, respectively. With $\mathcal{L}_{avqa}$ and the proposed two loss functions ($\mathcal{L}_{rmmr}$ and $\mathcal{L}_{ave}$), our AVQA framework can provide more accurate answers to questions even in the missing modality situations.

\section{Experimental Results}

\subsection{Experimental Settings}
\subsubsection{Dataset.} 
In the experiments, two publicly available open-source datasets, namely MUSIC-AVQA\cite{Music-AVQA} and AVQA\cite{2022avqa}, were utilized. The MUSIC-AVQA dataset serves as an AVQA benchmark for comprehensive scene understanding in musical performance. It encompasses 45,867 question-answer pairs derived from 9,288 videos. To elaborate further, 32,087, 4,595, and 9,185 question-answer pairs were allocated for training, validation, and testing, respectively.
Similarly, the AVQA dataset is a large-scale AVQA dataset designed for reasoning about multiple audio-visual relationships in real-life scenarios. It comprises 57,335 question-answer pairs sourced from 57,015 videos. In accordance with the specifications provided in \cite{2022avqa}, 34,401, 5,734, and 17,200 question-answer pairs were earmarked for training, validation, and testing, respectively.
\vspace{-0.3cm}
\subsubsection{AVQA Network.} To verify the capability of the proposed method, we adopted our method to four recently introduced AVQA networks (AVST \cite{Music-AVQA}\footnote[1]{\href{https://github.com/GeWu-Lab/MUSIC-AVQA}{https://github.com/GeWu-Lab/MUSIC-AVQA}}, AVSD \cite{AVSD}\footnote[2]{\href{https://github.com/idansc/simple-avsd}{https://github.com/idansc/simple-avsd}}, Pano-AVQA \cite{Pano-AVQA}\footnote[3]{\href{https://github.com/HS-YN/PanoAVQA}{https://github.com/HS-YN/PanoAVQA}}, and PSTP-Net \cite{PSTP}\footnote[4]{\href{https://github.com/GeWu-Lab/PSTP-Net}{https://github.com/GeWu-Lab/PSTP-Net}}) for which official code is available. Note that, all our implements are conducted by referring to the official codes. In our study, we selected AVST\cite{Music-AVQA} as our baseline, given its status as the latest network to exclusively utilize audio, visual, and text inputs. Given that PSTP-Net\cite{PSTP} integrates additional clip\cite{clip} features alongside these three modalities, it was not considered for our baseline. As such, AVST was used for conducting our ablation study experiments, being acknowledged as the most contemporary approach, except PSTP-Net.
\vspace{-0.3cm}

\subsubsection{Implementation Details.}
We train our AVQA framework on a single RTX 4090 GPU with a batch size of 4, utilizing the Adam optimizer \cite{ADAM} with an initial learning rate of $10^{-4}$. For the RMM Generator, we use $L=75$, and the default number of timesteps for the forward and reverse process of diffusion is 10. In our experiments, we set $\lambda_1=\lambda_2=1$.
\vspace{-0.3cm}

\begin{table*}[t!]
	\centering
	\begin{center}
		\renewcommand{\tabcolsep}{0.7mm}
		\caption{Results on MUSIC-AVQA dataset under missing modalities (visual missing (upper), audio missing (lower)) ($\mathcal{V}$: visual, $\mathcal{A}$: audio, $\mathcal{Q}$: question). $^{\ast}$ denotes a network that needs additional information (\textit{i.e.,} CLIP) for AVQA task.}
        \vspace{-0.3cm}
		\resizebox{0.999\linewidth}{!}
		{
		  \begin{tabular}{c | c | cc | c | cc | c | ccccc | c | c}
                \Xhline{3\arrayrulewidth}
                \rule{0pt}{10pt} \multirow{2}{*}[-0.5ex]{\bf Method} & \multirow{2}{*}[-0.5ex]{\bf Scenario} & \multicolumn{3}{c|}{\textbf{Audio Question}} & \multicolumn{3}{c|}{\textbf{Visual Question}} & \multicolumn{6}{c|}{\textbf{Audio-Visual Question}} & \multirow{2}{*}{\bf\makecell{All\\Avg}}\\\cline{3-5}\cline{6-8}\cline{9-14}
                & & \bf Cnt. & \bf Comp & \bf Avg & \bf Cnt. & \bf Loc & \bf Avg & \bf Exist & \bf Loc & \bf Cnt. & \bf Comp & \bf Temp & \bf Avg & \\\hline

                \rule{0pt}{10pt}
                \bf AVSD \cite{AVSD}       & \multirow{8}{*}{\makecell{Input: $\mathcal{A}+\mathcal{Q}$\\($\mathcal{V}$: Missing)}}
                                     & 68.10 & 61.56 & 65.68 & 59.22 & 55.54 & 57.35 & 72.36 & 56.51 & 52.10 & 57.76 & 50.61 & 58.11 & 59.25 \\
                \cellcolor{gray!20} \bf AVSD+Ours      &    & \cellcolor{gray!20}\bf 80.33& \cellcolor{gray!20}\bf 64.31 & \cellcolor{gray!20}\bf 74.43 & \cellcolor{gray!20}\bf 74.10 & \cellcolor{gray!20}\bf 64.41 & \cellcolor{gray!20}\bf 69.20 & \cellcolor{gray!20}\bf 79.25 & \cellcolor{gray!20}\bf 70.91 & \cellcolor{gray!20}\bf 57.83 & \cellcolor{gray!20}\bf 65.40 & \cellcolor{gray!20}\bf 58.88 & \cellcolor{gray!20}\bf 67.03 & \cellcolor{gray!20}\bf 68.91\\\cline{1-1} \cline{3-15}
                
                \rule{0pt}{10pt}
                \bf Pano-AVQA \cite{Pano-AVQA}  &   & 65.66 & 48.92 & 59.46 & 51.79 & 51.47 & 51.63 & 49.95 & 53.14 & 46.93 & 52.65 & 34.34 & 48.28 & 51.14 \\
                \cellcolor{gray!20}
                \bf Pano-AVQA+Ours &   & \cellcolor{gray!20}\bf 79.15 & \cellcolor{gray!20}\bf 64.14 & \cellcolor{gray!20}\bf 73.62 & \cellcolor{gray!20}\bf 72.51 & \cellcolor{gray!20}\bf 62.20 & \cellcolor{gray!20}\bf 67.30 & \cellcolor{gray!20}\bf 79.15 & \cellcolor{gray!20}\bf 70.51 & \cellcolor{gray!20}\bf 59.57 & \cellcolor{gray!20}\bf 63.49 & \cellcolor{gray!20}\bf 55.84 & \cellcolor{gray!20}\bf 66.33 & \cellcolor{gray!20}\bf 67.87 \\\cline{1-1} \cline{3-15}

                \rule{0pt}{10pt}
                \bf AVST\cite{Music-AVQA}       &  & 68.10 & 61.23 & 65.56 & 61.80 & 51.22 & 56.45 & \bf 80.70 & 55.10 & 48.00 & 60.09 & 45.87 & 58.38 & 59.14 \\
                \cellcolor{gray!20}
                \bf AVST+Ours      &  & \cellcolor{gray!20}\bf 78.27 & \cellcolor{gray!20}\bf 67.17 & \cellcolor{gray!20}\bf 74.18 & \cellcolor{gray!20}\bf 72.10 & \cellcolor{gray!20}\bf 64.08 & \cellcolor{gray!20}\bf 68.04 & \cellcolor{gray!20}77.94 & \cellcolor{gray!20}\bf 67.43 & \cellcolor{gray!20}\bf 58.48 & \cellcolor{gray!20}\bf 65.21 & \cellcolor{gray!20}\bf 58.88 & \cellcolor{gray!20}\bf 65.99 & \cellcolor{gray!20}\bf 67.98 \\\cline{1-1} \cline{3-15}
                
                \rule{0pt}{10pt}
                \bf PSTP-Net$^{\ast}$ \cite{PSTP}  &   & 70.40 & 62.79 & 67.60 & 58.06 & 51.59 & 54.79 & 80.16 & 55.81 & 46.74 & 58.58 & 51.34 & 58.77 & 59.27 \\
                \cellcolor{gray!20}
                \bf PSTP-Net$^{\ast}$+Ours &   & \cellcolor{gray!20}\bf 76.60 & \cellcolor{gray!20}\bf 65.49  & \cellcolor{gray!20}\bf 72.50  & \cellcolor{gray!20}\bf 68.50 & \cellcolor{gray!20}\bf 62.86 & \cellcolor{gray!20}\bf 65.65 & \cellcolor{gray!20}\bf 82.79 & \cellcolor{gray!20}\bf 63.95 & \cellcolor{gray!20}\bf 57.50 & \cellcolor{gray!20}\bf 61.13 & \cellcolor{gray!20}\bf 57.66 & \cellcolor{gray!20}\bf 64.82 & \cellcolor{gray!20}\bf 66.39 \\
                \hline\hline

                \rule{0pt}{10pt}
                \bf AVSD \cite{AVSD}       & \multirow{8}{*}{\makecell{Input: $\mathcal{V}+\mathcal{Q}$\\($\mathcal{A}$: Missing)}}
                                     & 38.45 & 55.74 & 44.86 & 42.20 & 37.30 & 39.72 & 58.09 & 44.74 & 27.08 & 49.78 & 15.41 & 40.53 & 41.08 \\
                \cellcolor{gray!20} \bf AVSD+Ours      &    & \cellcolor{gray!20}\bf 79.84 & \cellcolor{gray!20}\bf 64.81 & \cellcolor{gray!20}\bf 74.30 & \cellcolor{gray!20}\bf 74.94 & \cellcolor{gray!20}\bf 69.39 & \cellcolor{gray!20}\bf 72.13 & \cellcolor{gray!20}\bf 79.25 & \cellcolor{gray!20}\bf 71.15 & \cellcolor{gray!20}\bf 59.02 & \cellcolor{gray!20}\bf 65.03 & \cellcolor{gray!20}\bf 60.22 &  \cellcolor{gray!20}\bf 67.45 & \cellcolor{gray!20}\bf 69.90 \\\cline{1-1} \cline{3-15}
                
                \rule{0pt}{10pt}
                \bf Pano-AVQA \cite{Pano-AVQA}  &   & 39.24 & 57.57 & 46.03 & 43.45 & 35.50 & 39.43 & \bf80.10 & 39.17 & 26.43 & 49.87 & 16.99 & 43.56 & 42.91 \\
                \cellcolor{gray!20}
                \bf Pano-AVQA+Ours &   & \cellcolor{gray!20}\bf 79.45 & \cellcolor{gray!20}\bf 64.48 & \cellcolor{gray!20}\bf 73.93 & \cellcolor{gray!20}\bf 74.85 & \cellcolor{gray!20}\bf 69.80 & \cellcolor{gray!20}\bf72.30 & \cellcolor{gray!20}77.02 & \cellcolor{gray!20}\bf 71.46 & \cellcolor{gray!20}\bf 58.59 & \cellcolor{gray!20}\bf 64.21 & \cellcolor{gray!20}\bf 60.95 & \cellcolor{gray!20}\bf 69.95 & \cellcolor{gray!20}\bf 69.90 \\\cline{1-1} \cline{3-15}
                
                \rule{0pt}{10pt}
                \bf AVST \cite{Music-AVQA}       &  & 29.55 & 54.41 & 38.76 & 35.03 & 27.61 & 31.27 & 58.29 & 41.37 & 21.90 & 49.60 & 13.47 & 38.44 & 36.60 \\
                \cellcolor{gray!20}
                \bf AVST+Ours      &  & \cellcolor{gray!20}\bf 79.84 & \cellcolor{gray!20}\bf 64.81 & \cellcolor{gray!20}\bf 74.30 & \cellcolor{gray!20}\bf 74.77 & \cellcolor{gray!20}\bf 72.24 & \cellcolor{gray!20}\bf 73.49 & \cellcolor{gray!20}\bf 78.74 & \cellcolor{gray!20}\bf 70.59 & \cellcolor{gray!20}\bf 58.15 & \cellcolor{gray!20}\bf 62.67 & \cellcolor{gray!20}\bf 60.46 & \cellcolor{gray!20}\bf 66.44 & \cellcolor{gray!20}\bf 69.71 \\ 
                \cline{1-1} \cline{3-15}
                
                \rule{0pt}{10pt}
                \bf PSTP-Net$^{\ast}$\cite{PSTP}  &   & 78.76 & 55.72 & 70.27 & 75.52 & 71.35 & 73.41 & 77.73 & 68.54 & 54.67 & 59.85 & 58.03 & 64.25 & 67.74 \\
                \cellcolor{gray!20}
                \bf PSTP-Net$^{\ast}$+Ours &   & \cellcolor{gray!20}\bf 81.02 & \cellcolor{gray!20}\bf 60.94  & \cellcolor{gray!20}\bf 73.62  & \cellcolor{gray!20}\bf 78.20 & \cellcolor{gray!20}\bf 77.47 & \cellcolor{gray!20}\bf 77.83 & \cellcolor{gray!20}\bf 80.16 & \cellcolor{gray!20}\bf 71.38 & \cellcolor{gray!20}\bf 61.63 & \cellcolor{gray!20}\bf 62.03 & \cellcolor{gray!20}\bf 62.77 & \cellcolor{gray!20}\bf 67.92 & \cellcolor{gray!20}\bf 71.55 \\
                \specialrule{.1em}{.0em}{-.1em}
            \end{tabular}
		}
        \vspace{-0.5cm}
		\label{table:1}
	\end{center}
\end{table*}

%

\subsection{Evaluation Under Missing Modality.}

\noindent\textbf{Results on MUSIC-AVQA Dataset.} We adopt the four state-of-the-art AVQA networks, \textit{i.e.}, AVSD \cite{AVSD}, Pano-AVQA \cite{Pano-AVQA}, AVST \cite{Music-AVQA}, and PSTP-Net \cite{PSTP} on MUSIC-AVQA dataset \cite{Music-AVQA} to demonstrate the ability of our method in handling missing modalities. As shown in Table \ref{table:1}, when the visual modality is missing, existing AVQA methods struggle to estimate answers, achieving around 51$\sim$59\%. On the other hand, applying our approach to the existing AVQA networks significantly improves accuracy. Furthermore, our proposed method exhibits even more substantial improvements when the audio modality is missing. The results verify the effectiveness of our method in the missing modalities. \\

\noindent\textbf{Results on AVQA Dataset.} Furthermore, we extended our experiments to the AVQA dataset \cite{2022avqa}. Table \ref{tab:avqa} shows the results on the AVQA dataset. Notably, our method remains effective even when one modality is missed. These results demonstrate the efficacy of our proposed approach in compensating for missing modalities. Furthermore, our method exhibits flexibility enough, as it can seamlessly integrate into various existing AVQA network architectures.
\vspace{-0.3cm}

\subsection{Comparison with Existing Missing Modality Handling Methods.} 
We compare our method with the state-of-the-art methods \cite{missing_AAAI,missing_prompt,Shaspec,yao2024drfuse} that handle the missing modality on the MUSIC-AVQA dataset. We adopt AVST for the base AVQA backbone. As shown in Table \ref{table:3}, even in the visual modality missing and the audio modality missing, our method achieves the highest performance in overall accuracy (\textit{i.e.,} `All Avg' metric). The results show that even in the missing scenario, our RMM generator effectively recall the missing modal information. Also our AVR diffusion further enhances the feature representation of the audio-visual modality by leveraging the cross-modal relation.


\begin{table}[t]
\centering
\caption{Results on AVQA dataset under missing modalities (visual missing (upper), audio missing (lower)) ($\mathcal{V}$: visual, $\mathcal{A}$: audio, $\mathcal{Q}$: question).}\vspace{-0.3cm}
\resizebox{0.99\textwidth}{!}
{
\begin{tabular}{>{\centering\arraybackslash}p{4cm}|>{\centering\arraybackslash}p{2.5cm}|>{\centering\arraybackslash}p{1.5cm}>{\centering\arraybackslash}p{1.5cm}>{\centering\arraybackslash}p{1.5cm}>{\centering\arraybackslash}p{1.5cm}>{\centering\arraybackslash}p{1.5cm}>{\centering\arraybackslash}p{1.5cm}>{\centering\arraybackslash}p{1.5cm}>{\centering\arraybackslash}p{1.5cm}|c}
\Xhline{3\arrayrulewidth}
                                                &                                                                                      & \multicolumn{8}{c|}{\textbf{Question Type}}                                                                                                                                                                                                                                                                                           &                                                                              \\ \cline{3-10}
\multirow{-2}{*}{\textbf{Method}}               & \multirow{-2}{*}{\textbf{Scenario}}                                                  & \textbf{Used}                      & \textbf{When}                          & \textbf{\begin{tabular}[c]{@{}c@{}}Before\\ Next\end{tabular}}                   & \textbf{Why}                           & \textbf{Where}                         & \textbf{Happen}                     & \textbf{\begin{tabular}[c]{@{}c@{}}Come\\ From\end{tabular}}                     & \textbf{Which}                         & \multirow{-2}{*}{\textbf{\begin{tabular}[c]{@{}c@{}}All\\ Avg\end{tabular}}} \\ \hline
\textbf{AVSD \cite{AVSD}}                                   &                                                                                      & \bf52.94                                  & 42.86                                  & 56.00                                  & \bf 56.98                                  & 42.39                                  & 59.53                                  & \bf 44.65                                  & 40.36                                  & 45.80                                                                         \\
\cellcolor{gray!20}\textbf{AVSD+Ours}      &                                                                                      & \cellcolor{gray!20}39.63    & \cellcolor{gray!20}\textbf{45.45}    & \cellcolor{gray!20}\textbf{59.41}    & \cellcolor{gray!20}47.11    & \cellcolor{gray!20}\textbf{58.14}    & \cellcolor{gray!20}\textbf{68.00}    & \cellcolor{gray!20}42.86    & \cellcolor{gray!20}\textbf{64.71}    & \cellcolor{gray!20}\textbf{46.35}                                          \\ \cline{1-1} \cline{3-11} 
\textbf{Pano-AVQA \cite{Pano-AVQA}}                              &                                                                                      & 41.18                                  & 28.57                                  & 58.00                                  & 52.33                                  & 27.52                                  & 45.12                                  & 34.40                                  & 30.35                                  & 34.31                                                                        \\
\cellcolor{gray!20}\textbf{Pano-AVQA+Ours} &                                                                                      & \cellcolor{gray!20}\textbf{58.82} & \cellcolor{gray!20}\bf42.86 & \cellcolor{gray!20}\textbf{66.00} & \cellcolor{gray!20}\textbf{52.33} & \cellcolor{gray!20}\textbf{45.65} & \cellcolor{gray!20}\textbf{59.82} & \cellcolor{gray!20}\textbf{44.55} & \cellcolor{gray!20}\textbf{39.58} & \cellcolor{gray!20}\textbf{45.94}                                       \\ \cline{1-1} \cline{3-11} 
\textbf{AVST \cite{Music-AVQA}}                                   &                                                                                      & 35.29                                  & 23.81                                  & 56.00                                  & \bf54.65                                  & 27.95                                  & 41.07                                  & 33.98                                  & 28.83                                  & 32.84                                                                        \\
\cellcolor{gray!20}\textbf{AVST+Ours}      & \multirow{-6}{*}{\begin{tabular}[c]{@{}c@{}}Input: $\mathcal{A}+\mathcal{Q}$\\($\mathcal{V}$: Missing)\end{tabular}} & \cellcolor{gray!20}\textbf{58.82} & \cellcolor{gray!20}\textbf{42.86} & \cellcolor{gray!20}\textbf{64.00} & \cellcolor{gray!20}47.67 & \cellcolor{gray!20}\textbf{46.09} & \cellcolor{gray!20}\textbf{59.11} & \cellcolor{gray!20}\textbf{46.16} & \cellcolor{gray!20}\textbf{40.56} & \cellcolor{gray!20}\textbf{46.65}                                       \\ \hline \hline
\textbf{AVSD \cite{AVSD}}                                  &                                                                                      & \bf52.94                                  & 42.86                                  & 56.00                                  & \bf55.81                                  & 42.49                                  & 59.50                                  & \bf44.63                                  & 40.30                                  & 45.77                                                                        \\
\cellcolor{gray!20}\textbf{AVSD+Ours}      &                                                                                      & \cellcolor{gray!20}39.63    & \cellcolor{gray!20}\textbf{45.45}    & \cellcolor{gray!20}\textbf{59.41}    & \cellcolor{gray!20}47.11    & \cellcolor{gray!20}\textbf{58.14}    & \cellcolor{gray!20}\textbf{68.00}    & \cellcolor{gray!20}42.86    & \cellcolor{gray!20}\textbf{64.71}    & \cellcolor{gray!20}\textbf{46.38}                                           \\ \cline{1-1} \cline{3-11} 
\textbf{Pano-AVQA \cite{Pano-AVQA}}                              &                                                                                      & \bf70.59                                  & 52.38                                  & 68.00                                  & \bf58.14                                  & 53.77                                  & 64.64                                  & 57.60                                  & 51.71                                  & 56.32                                                                        \\
\cellcolor{gray!20}\textbf{Pano-AVQA+Ours} &                                                                                      & \cellcolor{gray!20}64.71 & \cellcolor{gray!20}\textbf{61.90} & \cellcolor{gray!20}\textbf{74.00} & \cellcolor{gray!20}55.81 & \cellcolor{gray!20}\textbf{72.44} & \cellcolor{gray!20}\textbf{71.78} & \cellcolor{gray!20}\textbf{72.35} & \cellcolor{gray!20}\textbf{70.08} & \cellcolor{gray!20}\textbf{71.28}                                       \\ \cline{1-1} \cline{3-11} 
\textbf{AVST \cite{Music-AVQA}}                                   &                                                                                      & 52.94                                  & 47.62                                  & 62.00                                  & 56.98                                  & 55.08                                  & 65.06                                  & 58.58                                  & 52.39                                  & 57.03                                                                        \\
\cellcolor{gray!20}\textbf{AVST+Ours}      & \multirow{-6}{*}{\begin{tabular}[c]{@{}c@{}}Input: $\mathcal{V}+\mathcal{Q}$\\($\mathcal{A}$: Missing)\end{tabular}} & \cellcolor{gray!20}\textbf{70.59} & \cellcolor{gray!20}\textbf{57.14} & \cellcolor{gray!20}\textbf{70.00} & \cellcolor{gray!20}\textbf{59.30} & \cellcolor{gray!20}\textbf{72.34} & \cellcolor{gray!20}\textbf{72.96} & \cellcolor{gray!20}\textbf{73.77} & \cellcolor{gray!20}\textbf{65.86} & \cellcolor{gray!20}\textbf{70.28}                                       \\ \specialrule{.1em}{.0em}{-.1em}
\end{tabular}}
\label{tab:avqa}
\end{table}

\begin{table*}[t!]
	\centering
	\begin{center}
		\renewcommand{\tabcolsep}{0.4mm}
		\caption{Comparison of our method with recent approaches for handling missing modalities in the MUSIC-AVQA dataset (visual missing (upper), audio missing (lower)) ($\mathcal{V}$: visual, $\mathcal{A}$: audio, $\mathcal{Q}$: question). We adopt AVST, denoted as $\mathcal{B}$, for the baseline of AVQA task. \textbf{Bold}/\underline{underlined} fonts indicate the best/second-best results.}
        \vspace{-0.3cm}
		\resizebox{0.999\linewidth}{!}
		{
		  \begin{tabular}{c | c | cc | c | cc | c | ccccc | c | c}
                \Xhline{3\arrayrulewidth}
                \rule{0pt}{10pt} \multirow{2}{*}[-0.5ex]{\bf Method} & \multirow{2}{*}[-0.5ex]{\bf Scenario} & \multicolumn{3}{c|}{\textbf{Audio Question}} & \multicolumn{3}{c|}{\textbf{Visual Question}} & \multicolumn{6}{c|}{\textbf{Audio-Visual Question}} & \multirow{2}{*}{\bf\makecell{All\\Avg}}\\\cline{3-5}\cline{6-8}\cline{9-14}
                & & \bf Cnt. & \bf Comp & \bf Avg & \bf Cnt. & \bf Loc & \bf Avg & \bf Exist & \bf Loc & \bf Cnt. & \bf Comp & \bf Temp & \bf Avg & \\\hline

                                \rule{0pt}{10pt}
                AVST ($\mathcal{B}$) \cite{Music-AVQA} & \multirow{5}{*}{\makecell{Input: $\mathcal{V}+\mathcal{Q}$\\($\mathcal{A}$: Missing)}}
                              & 29.55 & 54.41 & 38.76 & 35.03 & 27.61 & 31.27 & 58.29 & 41.37 & 21.90 & 49.60 & 13.47 & 38.44 & 36.60 \\\cdashline{1-1}\cdashline{3-15}
                \rule{0pt}{10pt}
                            
                $\mathcal{B}$+Lee et al. \cite{missing_prompt} (CVPR'23)
                            & & 69.91 & 63.47 & 67.54 & 58.40 & 55.35 & 56.85 & \bf 80.67 & 57.23 & 47.93 & \underline{62.03} & 47.81 & 59.62 & 60.28 \\

                $\mathcal{B}$+ShaSpec\cite{Shaspec}(CVPR'23)
                            & & 76.99 & 59.76 & 70.64 & 72.35 & 66.69 & 69.49 & \underline{79.15} & 66.09 & 53.26 & 61.22 & 56.20 & 63.66 & 66.44 \\

                $\mathcal{B}$+Woo et al. \cite{missing_AAAI} (AAAI'23)
                            & & \underline{77.29} & \underline{63.64} & \underline{72.25} & 72.26 & \underline{67.59} & \underline{69.90} & 79.05 & \underline{68.06} & \underline{55.76} & 61.13 & \underline{56.57} & \underline{64.62} & \underline{67.37} \\
                            
                $\mathcal{B}$+Yao et al. \cite{yao2024drfuse} (AAAI'24)
                            & & 77.09 & 59.43 & 70.58 & \underline{72.43} & 67.10 & 69.74 & \underline{79.15} & 65.69 & 53.04 & 61.94 & 56.20 & 63.68 & 66.50 \\
                            
                \cellcolor{gray!20}\bf $\mathcal{B}$+Ours      &  & 
                \cellcolor{gray!20}\bf 79.84 & \cellcolor{gray!20}\bf 64.81 & \cellcolor{gray!20}\bf 74.30 & \cellcolor{gray!20}\bf 74.77 & \cellcolor{gray!20}\bf 72.24 & \cellcolor{gray!20}\bf 73.49 & \cellcolor{gray!20}78.74 & \cellcolor{gray!20}\bf 70.59 & \cellcolor{gray!20}\bf 58.15 & \cellcolor{gray!20}\bf62.67 & \cellcolor{gray!20}\bf 60.46 & \cellcolor{gray!20}\bf 66.44 & \cellcolor{gray!20}\bf 69.71 \\             
                \hline\hline

                \rule{0pt}{10pt}
                AVST ($\mathcal{B}$) \cite{Music-AVQA} & \multirow{5}{*}{\makecell{Input: $\mathcal{A}+\mathcal{Q}$\\($\mathcal{V}$: Missing)}}
                            & 68.10 & 61.23 & 65.56 & 61.80 & 51.22 & 56.45 & \underline{80.70} & 55.10 & 48.00 & 60.09 & 45.87 & 58.38 & 59.14 \\\cdashline{1-1}\cdashline{3-15}
                \rule{0pt}{10pt}
                            
                $\mathcal{B}$+Lee et al. \cite{missing_prompt} (CVPR'23) 
                            & & 71.39 & 64.14 & 68.72 & 63.32 & 59.18 & 61.23 & \bf 81.38 & 60.32 & 53.59 & 61.04 & 54.62 & 62.42 & 63.22 \\
  
                $\mathcal{B}$+ShaSpec\cite{Shaspec} (CVPR'23) 
                            & & 77.58 & \bf67.51 & 73.87 & \textbf{72.43} & 64.00 & 68.17 & 78.04 & 66.80 & 57.83 & 60.67 & 55.96 & 64.29 & 67.01 \\

                $\mathcal{B}$+Woo et al. \cite{missing_AAAI} (AAAI'23)
                            & & \bf78.66 & 65.82 & \underline{73.93} & 71.60 & \bf65.31 & \underline{68.41} & 78.54 & 65.77 & 57.39 & \underline{61.58} & 56.20 & 64.29 & 67.08  \\
                            
                $\mathcal{B}$+Yao et al. \cite{yao2024drfuse} (AAAI'24)
                            & & 77.78 & \underline{67.17} & 73.87 & \underline{72.51} & \underline{64.73} & \bf 68.58 & 77.83 & \underline{66.96} & \bf 58.59 & 59.49 & \underline{57.42} & \underline{64.40} & \underline{67.18} \\
                            
                \cellcolor{gray!20}
                \bf $\mathcal{B}$+Ours      &  & \cellcolor{gray!20}\underline{78.27} & \cellcolor{gray!20}\underline{67.17} & \cellcolor{gray!20}\bf74.18 & \cellcolor{gray!20} 72.10 & \cellcolor{gray!20}64.08 & \cellcolor{gray!20} 68.04 & \cellcolor{gray!20}77.94 & \cellcolor{gray!20}\bf 67.43 & \cellcolor{gray!20}\underline{58.48} & \cellcolor{gray!20}\bf 65.21 & \cellcolor{gray!20}\bf 58.88 & \cellcolor{gray!20}\bf 65.99 & \cellcolor{gray!20}\bf 67.98 \\ \specialrule{.1em}{.0em}{-.1em}
            \end{tabular}
		}
            \vspace{-0.5cm}
		\label{table:3}
	\end{center}
\end{table*}


\begin{table}[t]
\caption{Effect of pseudo features generated from each component with missing modalities, by adopting AVST model on MUSIC-AVQA dataset. RMM denotes RMM Generator and AVR refers to AVR diffusion model ($\mathcal{V}$: visual, $\mathcal{A}$: audio, $\mathcal{Q}$: question).}
\vspace{-0.3cm}
\resizebox{0.99\linewidth}{!}{
\begin{tabular}{c|cc|ccc|c}
\specialrule{.1em}{.0em}{-.1em}
                                                                                    & \multicolumn{2}{c|}{\textbf{Components}}              & \multicolumn{3}{c|}{\textbf{Question Type}}                                                                              &                                                                              \\ \cline{2-6}
\multirow{-2}{*}{\textbf{Scenario}}                                                 & \textbf{RMM}           & \textbf{AVR}        & \textbf{Audio Question}                & \textbf{Visual Question}               & \textbf{Audio-Visual Question}         & \multirow{-2}{*}{\textbf{\begin{tabular}[c]{@{}c@{}}All\\ Avg\end{tabular}}} \\ \hline
                                                                                    & \xmark                         & \xmark                         & 65.56                                  & 56.45                                  & 58.38                                  & 59.14                                                                        \\
                                                                                    & \cmark                         & \xmark                         & 69.50                                  & 59.74                                  & 62.22                                  & 62.85                                                                        \\
                                                                                    & \xmark                         & \cmark                         & 73.00                                  & 66.23                                  & 62.99                                  & 65.62                                                                        \\
\multirow{-4}{*}{\begin{tabular}[c]{@{}c@{}}Input: $\mathcal{A}+\mathcal{Q}$\\ ($\mathcal{V}$: Missing)\end{tabular}} & \cellcolor{gray!20}\cmark & \cellcolor{gray!20}\cmark & \cellcolor{gray!20}\textbf{74.18} & \cellcolor{gray!20}\textbf{68.04} & \cellcolor{gray!20}\textbf{65.99} & \cellcolor{gray!20}\textbf{67.98}                                       \\ \hline \hline
                                                                                    & \xmark                         & \xmark                         & 38.76                                  & 31.27                                  & 38.44                                  & 36.60                                                                         \\
                                                                                    & \cmark                         & \xmark                         & 68.64                                  & 57.93                                  & 58.93                                  & 60.38                                                                        \\
                                                                                    & \xmark                         & \cmark                         & 69.71                                  & 70.02                                  & 62.97                                  & 66.03                                                                        \\
\multirow{-4}{*}{\begin{tabular}[c]{@{}c@{}}Input: $\mathcal{V}+\mathcal{Q}$\\ ($\mathcal{A}$: Missing)\end{tabular}} & \cellcolor{gray!20}\cmark & \cellcolor{gray!20}\cmark & \cellcolor{gray!20}\textbf{74.30} & \cellcolor{gray!20}\textbf{73.49} & \cellcolor{gray!20}\textbf{66.44} & \cellcolor{gray!20}\textbf{69.71}                                       \\ \specialrule{.1em}{.0em}{-.1em}
\end{tabular}
}
\label{tab:ablation}
\vspace{-0.2cm}
\end{table}

\begin{table}[t!]
\centering
\begin{minipage}[t]{0.49\textwidth}
    \centering
    \caption{AVQA results on MUSIC-AVQA by changing the number of slots.}
    \vspace{-0.3cm}
    \resizebox{0.8\textwidth}{!}{
    \begin{tabularx}{\textwidth}{>{\centering\arraybackslash}X|>{\centering\arraybackslash}X|>{\centering\arraybackslash}X}
    \specialrule{.1em}{.0em}{-.05em}
    \multirow{2}{*}{\begin{tabular}[c]{@{}c@{}}\# of Slot\end{tabular}} & \multicolumn{2}{c}{Missing Modality} \\ 
    \cline{2-3}
    & $\mathcal{V}$ & $\mathcal{A}$ \\ 
    \hline
    - & 59.14 & 36.60 \\ \hdashline
    25 & 67.41 & 68.75 \\
    50 & 67.23 & 68.85 \\
    \textbf{75} & \textbf{67.98} & \textbf{69.71} \\
    100 & 67.17 & 68.85 \\ 
    \specialrule{.1em}{.0em}{.0em}
    \end{tabularx}}
    \label{tab:slot}
\end{minipage}\hfill
\begin{minipage}[t]{0.49\textwidth}
    \setlength{\extrarowheight}{1.5pt}
    \centering
    \caption{AVQA results on MUSIC-AVQA by changing time steps of diffusion process.}
    \vspace{-0.2cm}
    \resizebox{0.8\textwidth}{!}{
    \begin{tabularx}{\textwidth}{>{\centering\arraybackslash}X|>{\centering\arraybackslash}X|>{\centering\arraybackslash}X}
    \specialrule{.1em}{.0em}{.0em}
    \multirow{2}{*}{\begin{tabular}[c]{@{}c@{}}\# of\\ Time Step\end{tabular}} & \multicolumn{2}{c}{Missing Modality} \\ \cline{2-3} 
                                                                               & $\mathcal{V}$                 & $\mathcal{A}$                \\ \hline
    -                                                                          & 62.85             & 60.38            \\ \hdashline
    5                                                                          & 67.51             & 67.04            \\
    \textbf{10}                                                                         & \textbf{67.98}             & \textbf{69.71}            \\
    20                                                                         & 67.37             & 68.20            \\ \specialrule{.1em}{.0em}{.0em}
    \end{tabularx}}
    \label{tab:timestep}
\end{minipage}
\vspace{-0.5cm}
\end{table}
\subsection{Ablation Study}
\noindent\textbf{Effect of the RMM Generator and AVR Diffusion Model.}
We conducted experiments to evaluate the effectiveness of pseudo features generated by our proposed RMM generator and AVR diffusion model. Table \ref{tab:ablation} presents the results when features for the missing modality are generated by each module. The absence of the visual modality leads to a decline in accuracy, particularly for questions related to visual information. However, leveraging pseudo features generated by the RMM generator helps alleviate the impact of the missing modality, leading to improved accuracy. Furthermore, employing the diffusion model for pseudo feature generation yields even better results compared to the RMM. While the RMM generator relies on similarity-based feature generation, the diffusion model surpasses it by learning the generating more enhanced representations. We also utilized the features generated from the RMM generator as the initial input for the AVR diffusion model, enabling it to be effectively enhanced. Consequently, by integrating the RMM generator and AVR diffusion model, we generated enhanced features learned with cross-modal knowledge. \\

\noindent\textbf{Slot Number of RMM Generator Network.}
To investigate the effect of varying the number of slots in the RMM network, we ran experiments varying the slot size over a range of values: \{25, 50, 75, 100\}. Table \ref{tab:slot} shows the results of the ablation study corresponding to each slot number. As shown in the table, performance peaks when the slot number is set to 75, regardless of the missing modality scenarios. Furthermore, even with variations in the slot number, our method consistently outperforms the baseline approach across all settings. \\

\noindent\textbf{Effect of the Number of Time Steps in Diffusion Process.}
In this experiment, we investigate the impact of varying the number of time steps in the diffusion process. Specifically, we examine three different settings: 5, 10, and 20 time steps. As depicted in the Table \ref{tab:timestep}, the AVQA performance is optimal when the number of time steps is set to 10. Consequently, we adopt this value for all subsequent experiments.
\begin{figure}[t]
    \centering
        \centerline{\includegraphics[width=0.82\linewidth]{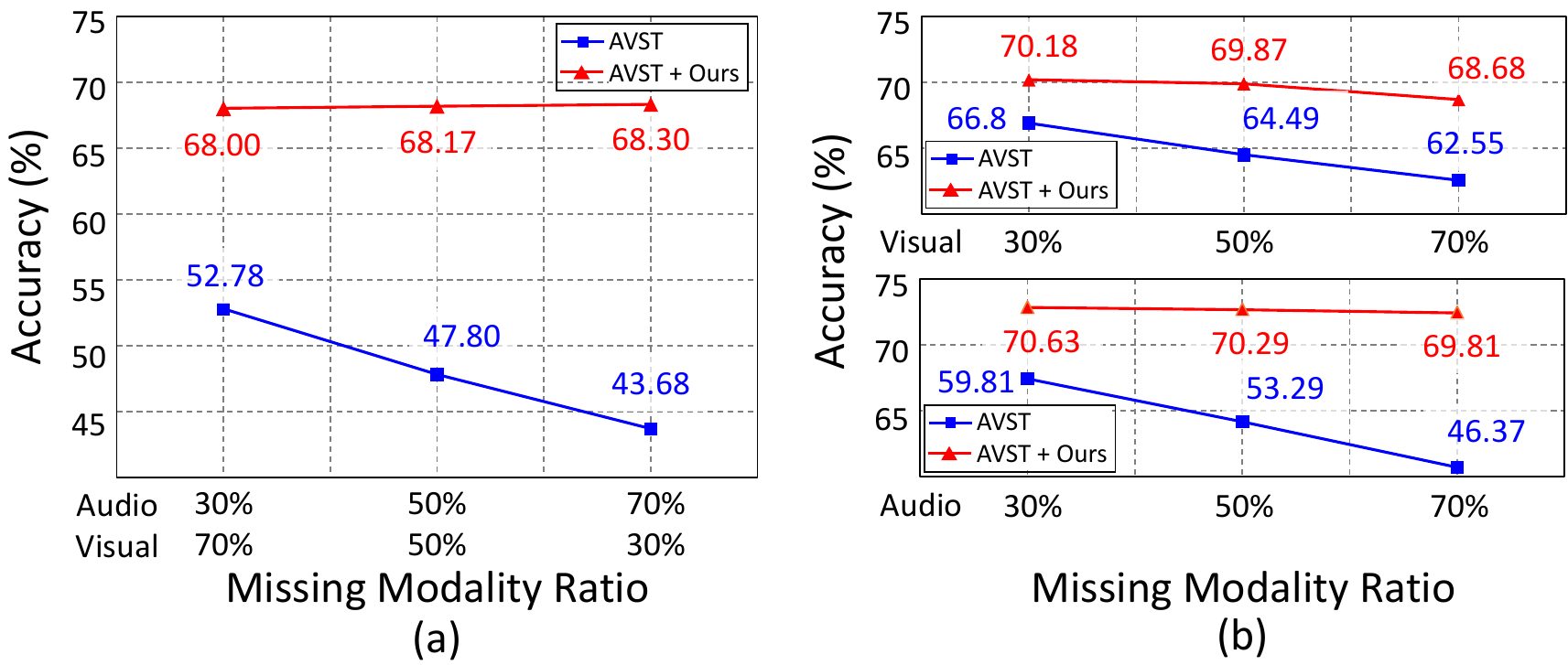}}
        \vspace{-0.2cm}
	\caption{AVQA results on the MUSIC-AVQA dataset vary based on (a) the missing ratio of visual and audio modalities, and (b) the varying missing ratios for visual (upper) or audio (lower) modalities.}
    \label{fig:5}
    \vspace{-0.4cm}
\end{figure}
\subsection{Discussions}
\noindent\textbf{Varying Missing Ratios.}
We also experimented with various missing rates to test aspects similar to real-world applications. In \cref{fig:5} (a), we can see that visual and audio perform consistently well even when one of them is missing with a higher ratio. In \cref{fig:5} (b), we can see that the probability of missing each modality is not significantly affected, and performs consistently well. \cref{fig:5} results prove that our method can work robustly in terms of real-world applications such as various missing ratio situations. \\

\noindent\textbf{Limitation.}
We addresses the problem of missing modalities in only inference situations that are executed after training has taken. However, in terms of further real-world applications, it is possible that missing modalities may occur during learning. So in future work, this consideration will lead to the study of AVQA networks that can robustly cope with missing modalities in training situations.

\section{Conclusion}

In this work, we introduced a novel Audio-Visual Question Answering (AVQA) framework designed to tackle the challenge of missing modalities in real-world scenarios. Our framework incorporates the Relation-aware Missing Modal (RMM) generator and the Audio-Visual Relation-aware (AVR) diffusion model. The RMM generator generates the pseudo feature of the missing modality, while the AVR diffusion model enhances audio-visual representations. It effectively handles situations where audio or visual information is missing. Through our experiments and comparisons with state-of-the-art AVQA methods, we demonstrated the superior performance of our approach, even in scenarios where one modality is missing. This contributes to enhancing the robustness and accuracy of AVQA networks in real-world environments.

\section*{Acknowledgements}
This work was supported by the NRF grant funded by the Korea government (MSIT) (No. RS-2023-00252391), and by the IITP grant funded by the Korea government (MSIT) (No. 2022-0-00124: Development of Artificial Intelligence Technology for Self-Improving Competency-Aware Learning Capabilities, No. RS-2022-00155911: Artificial Intelligence Convergence Innovation Human Resources Development (Kyung Hee University), IITP-2023-RS-2023-00266615: Convergence Security Core Talent Training Business Support Program), and conducted by CARAI grant funded by DAPA and ADD (UD230017TD).

%
%

\bibliographystyle{splncs04}

\begin{thebibliography}{10}
\providecommand{\url}[1]{\texttt{#1}}
\providecommand{\urlprefix}{URL }
\providecommand{\doi}[1]{https://doi.org/#1}

\bibitem{missing_adversarial}
Cai, L., Wang, Z., Gao, H., Shen, D., Ji, S.: Deep adversarial learning for multi-modality missing data completion. In: Int. Conf. Knowledge Discovery and Data Mining (2018)

\bibitem{calvert2001detection}
Calvert, G.A., Hansen, P.C., Iversen, S.D., Brammer, M.J.: Detection of audio-visual integration sites in humans by application of electrophysiological criteria to the bold effect. Neuroimage  (2001)

\bibitem{av2}
Chen, Y., Xian, Y., Koepke, A., Shan, Y., Akata, Z.: Distilling audio-visual knowledge by compositional contrastive learning. In: IEEE Conf. Comput. Vis. Pattern Recog. (2021)

\bibitem{driving1}
Choi, C., Choi, J.H., Li, J., Malla, S.: Shared cross-modal trajectory prediction for autonomous driving. In: IEEE Conf. Comput. Vis. Pattern Recog. (2021)

\bibitem{dhariwal2021diffusion}
Dhariwal, P., Nichol, A.: Diffusion models beat gans on image synthesis. Adv. Neural Inform. Process. Syst.  (2021)

\bibitem{gu2022vector}
Gu, S., Chen, D., Bao, J., Wen, F., Zhang, B., Chen, D., Yuan, L., Guo, B.: Vector quantized diffusion model for text-to-image synthesis. In: IEEE Conf. Comput. Vis. Pattern Recog. (2022)

\bibitem{DDPM}
Ho, J., Jain, A., Abbeel, P.: Denoising diffusion probabilistic models. Adv. Neural Inform. Process. Syst.  (2020)

\bibitem{missingmodal_decoding}
Jin, T., Cheng, X., Li, L., Lin, W., Wang, Y., Zhao, Z.: Rethinking missing modality learning from a decoding perspective. In: ACM Int. Conf. Multimedia (2023)

\bibitem{kawar2022denoising}
Kawar, B., Elad, M., Ermon, S., Song, J.: Denoising diffusion restoration models. Adv. Neural Inform. Process. Syst.  (2022)

\bibitem{NoFV}
Kim, D., Park, K., Lee, G.: Oddeyecam: A sensing technique for body-centric peephole interaction using wfov rgb and nfov depth cameras. In: ACM Symp. User Interface Software Technology (2020)

\bibitem{kim2024learning}
Kim, D., Um, S.J., Lee, S., Kim, J.U.: Learning to visually localize sound sources from mixtures without prior source knowledge. In: IEEE Conf. Comput. Vis. Pattern Recog. (2024)

\bibitem{kim2021diffusionclip}
Kim, G., Kwon, T., Ye, J.C.: Diffusionclip: Text-guided diffusion models for robust image manipulation. In: IEEE Conf. Comput. Vis. Pattern Recog. (2022)

\bibitem{kim2022towards}
Kim, J.U., Park, S., Ro, Y.M.: Towards versatile pedestrian detector with multisensory-matching and multispectral recalling memory. In: AAAI (2022)

\bibitem{kim2023enabling}
Kim, J.U., Ro, Y.M.: Enabling visual object detection with object sounds via visual modality recalling memory. IEEE Trans. Neural Netw. Learn. Syst.  (2023)

\bibitem{ADAM}
Kingma, D.P., Ba, J.: Adam: A method for stochastic optimization. arXiv preprint arXiv:1412.6980  (2014)

\bibitem{lahat2015multimodal}
Lahat, D., Adali, T., Jutten, C.: Multimodal data fusion: an overview of methods, challenges, and prospects. Proceedings of the IEEE  (2015)

\bibitem{lee2022weakly}
Lee, S., Kim, H.I., Ro, Y.M.: Weakly paired associative learning for sound and image representations via bimodal associative memory. In: IEEE Conf. Comput. Vis. Pattern Recog. (2022)

\bibitem{lee2022audio}
Lee, S., Park, S., Ro, Y.M.: Audio-visual mismatch-aware video retrieval via association and adjustment. In: Eur. Conf. Comput. Vis. (2022)

\bibitem{missing_prompt}
Lee, Y.L., Tsai, Y.H., Chiu, W.C., Lee, C.Y.: Multimodal prompting with missing modalities for visual recognition. In: IEEE Conf. Comput. Vis. Pattern Recog. (2023)

\bibitem{PSTP}
Li, G., Hou, W., Hu, D.: Progressive spatio-temporal perception for audio-visual question answering. In: ACM Int. Conf. Multimedia (2023)

\bibitem{Music-AVQA}
Li, G., Wei, Y., Tian, Y., Xu, C., Wen, J.R., Hu, D.: Learning to answer questions in dynamic audio-visual scenarios. In: IEEE Conf. Comput. Vis. Pattern Recog. (2022)

\bibitem{human1}
Lindenberger, U.: Human cognitive aging: corriger la fortune? science  (2014)

\bibitem{Ma_2022_CVPR}
Ma, M., Ren, J., Zhao, L., Testuggine, D., Peng, X.: Are multimodal transformers robust to missing modality? In: IEEE Conf. Comput. Vis. Pattern Recog. (2022)

\bibitem{missingmodal}
Ma, M., Ren, J., Zhao, L., Tulyakov, S., Wu, C., Peng, X.: Smil: Multimodal learning with severely missing modality. In: AAAI (2021)

\bibitem{missingmodal_ssm}
Maheshwari, H., Liu, Y.C., Kira, Z.: Missing modality robustness in semi-supervised multi-modal semantic segmentation. In: IEEE Winter Conf. Appl. Comput. Vis. (2024)

\bibitem{av3}
Majumder, S., Chen, C., Al-Halah, Z., Grauman, K.: Few-shot audio-visual learning of environment acoustics. Adv. Neural Inform. Process. Syst.  (2022)

\bibitem{human3}
McGrew, K.S.: Chc theory and the human cognitive abilities project: Standing on the shoulders of the giants of psychometric intelligence research (2009)

\bibitem{meng2021sdedit}
Meng, C., He, Y., Song, Y., Song, J., Wu, J., Zhu, J.Y., Ermon, S.: Sdedit: Guided image synthesis and editing with stochastic differential equations. arXiv preprint arXiv:2108.01073  (2021)

\bibitem{nichol2021glide}
Nichol, A., Dhariwal, P., Ramesh, A., Shyam, P., Mishkin, P., McGrew, B., Sutskever, I., Chen, M.: Glide: Towards photorealistic image generation and editing with text-guided diffusion models. arXiv preprint arXiv:2112.10741  (2021)

\bibitem{missingmodal_av}
Parthasarathy, S., Sundaram, S.: Training strategies to handle missing modalities for audio-visual expression recognition. In: Proc. ACM Int. Conf. Multimodal Interact. (2020)

\bibitem{av1}
Pian, W., Mo, S., Guo, Y., Tian, Y.: Audio-visual class-incremental learning. In: Int. Conf. Comput. Vis. (2023)

\bibitem{qiu2022learning}
Qiu, Z., Yang, H., Fu, J., Fu, D.: Learning spatiotemporal frequency-transformer for compressed video super-resolution. In: Eur. Conf. Comput. Vis. (2022)

\bibitem{clip}
Radford, A., Kim, J.W., Hallacy, C., Ramesh, A., Goh, G., Agarwal, S., Sastry, G., Askell, A., Mishkin, P., Clark, J., et~al.: Learning transferable visual models from natural language supervision. In: Int. Conf. Mach. Learn. (2021)

\bibitem{raij2000audiovisual}
Raij, T., Uutela, K., Hari, R.: Audiovisual integration of letters in the human brain. Neuron  (2000)

\bibitem{rombach2022high}
Rombach, R., Blattmann, A., Lorenz, D., Esser, P., Ommer, B.: High-resolution image synthesis with latent diffusion models. In: IEEE Conf. Comput. Vis. Pattern Recog. (2022)

\bibitem{saharia2022palette}
Saharia, C., Chan, W., Chang, H., Lee, C., Ho, J., Salimans, T., Fleet, D., Norouzi, M.: Palette: Image-to-image diffusion models. In: ACM SIGGRAPH (2022)

\bibitem{saharia2022image}
Saharia, C., Ho, J., Chan, W., Salimans, T., Fleet, D.J., Norouzi, M.: Image super-resolution via iterative refinement. IEEE Trans. Pattern Anal. Mach. Intell.  (2022)

\bibitem{AVSD}
Schwartz, I., Schwing, A.G., Hazan, T.: A simple baseline for audio-visual scene-aware dialog. In: IEEE Conf. Comput. Vis. Pattern Recog. (2019)

\bibitem{human2}
Sweller, J.: Instructional design consequences of an analogy between evolution by natural selection and human cognitive architecture. Instructional science  (2004)

\bibitem{um2023audio}
Um, S.J., Kim, D., Kim, J.U.: Audio-visual spatial integration and recursive attention for robust sound source localization. In: ACM Int. Conf. Multimedia (2023)

\bibitem{Shaspec}
Wang, H., Chen, Y., Ma, C., Avery, J., Hull, L., Carneiro, G.: Multi-modal learning with missing modality via shared-specific feature modelling. In: IEEE Conf. Comput. Vis. Pattern Recog. (2023)

\bibitem{missing_AAAI}
Woo, S., Lee, S., Park, Y., Nugroho, M.A., Kim, C.: Towards good practices for missing modality robust action recognition. In: AAAI (2023)

\bibitem{driving2}
Wu, R., Wang, H., Dayoub, F., Chen, H.T.: Segment beyond view: Handling partially missing modality for audio-visual semantic segmentation. In: AAAI (2024)

\bibitem{2022avqa}
Yang, P., Wang, X., Duan, X., Chen, H., Hou, R., Jin, C., Zhu, W.: Avqa: A dataset for audio-visual question answering on videos. In: ACM Int. Conf. Multimedia (2022)

\bibitem{yao2024drfuse}
Yao, W., Yin, K., Cheung, W.K., Liu, J., Qin, J.: Drfuse: Learning disentangled representation for clinical multi-modal fusion with missing modality and modal inconsistency. In: AAAI (2024)

\bibitem{Pano-AVQA}
Yun, H., Yu, Y., Yang, W., Lee, K., Kim, G.: Pano-avqa: Grounded audio-visual question answering on 360deg videos. In: Int. Conf. Comput. Vis. (2021)

\bibitem{zeng2021improving}
Zeng, Y., Yang, H., Chao, H., Wang, J., Fu, J.: Improving visual quality of image synthesis by a token-based generator with transformers. Adv. Neural Inform. Process. Syst.  (2021)

\bibitem{av4}
Zhang, J., Xu, X., Shen, F., Lu, H., Liu, X., Shen, H.T.: Enhancing audio-visual association with self-supervised curriculum learning. In: AAAI (2021)

\end{thebibliography}

\end{document}